\documentclass[runningheads]{llncs}
\usepackage[square,numbers]{natbib}

\usepackage[T1]{fontenc}
\usepackage{graphicx}
\usepackage{tabularx}
\usepackage{booktabs}
\usepackage{subfig}
\begin{document}
\title{Grounding Toxicity in Real-World Events across Languages}
%
%

\author{Wondimagegnhue Tsegaye Tufa \and
Ilia Markov \and
Piek Vossen }
\authorrunning{W. T. Tufa et al.}
\institute{Vrije Universiteit Amsterdam \\
De Boelelaan 1105, 1081 HV Amsterdam, The Netherlands \\
\{w.t.tufa, i.markov, p.t.j.m.vossen\}@vu.nl
}

%
\maketitle              
\begin{abstract}
Social media conversations frequently suffer from toxicity, creating significant issues for users, moderators, and entire communities. Events in the real world, like elections or conflicts, can initiate and escalate toxic behavior online. Our study investigates how real-world events influence the origin and spread of toxicity in online discussions across various languages and regions. We gathered Reddit data comprising 4.5 million comments from 31 thousand posts in six different languages (Dutch, English, German, Arabic, Turkish and Spanish). We target fifteen major social and political world events that occurred between 2020 and 2023. We observe significant variations in toxicity, negative sentiment, and emotion expressions across different events and language communities, showing that toxicity is a complex phenomenon in which many different factors interact and still need to be investigated. We will release the data for further research along with our code.

\end{abstract}
\section{Introduction}
Social media platforms have experienced significant growth in their user base and importance as communication tools. They provide a space where individuals can freely express various opinions. This openness and accessibility lead to a diverse mix of content, from insightful and valuable perspectives to controversial or even offensive content \cite{fortuna2018survey,mathew2021hatexplain}.

Significant real-world events, including GamerGate in August 2014, the murder of George Floyd in May 2020, and the January 6th insurrection at the US Capitol in 2021 have been key triggers for toxic reactions on platforms such as Reddit \cite{chatzakou2017measuring,kumar2023understanding}.
Such events, which are often filled with strong emotions and diverse opinions, frequently create conditions that foster conflicts and aggressive interactions, which can lead to divisive discourse, where individuals are more prone to express extreme views or engage in aggressive behavior \cite{hiaeshutter2022language,mall2020four}. The anonymity and absence of face-to-face communication provided by online platforms like Reddit further lower the barriers to expressing online toxicity.

The detection of toxic content has become an increasingly important research topic in Natural Language Processing (NLP). Current studies in this domain concentrate on developing datasets encompassing various dimensions of toxic content \cite{mathew2021hatexplain,vidgen2021introducing,sachdeva-etal-2022-measuring}, and training models that utilize these datasets to build systems for classifying and filtering toxic content \cite{schouten2023cross,gevers2022linguistic,radfar2020characterizing,van2018challenges}.

The prevalence of toxic language on platforms like Reddit has been widely researched. These studies focus on aspects such as user comments and posts~\cite{kumar2023understanding,hiaeshutter2022language}, community-level interactions \cite{farrell2019exploring,urbaniak2022namespotting}, or the behavior of the users \cite{urbaniak2022namespotting}. 


Inappropriate interactions like toxicity do not occur in a vacuum. Various factors can trigger toxic behavior in online discussions,  for instance, the specific topic being discussed, as well as the presence of threats or inflammatory comments \cite{hiaeshutter2022language,salminen2020topic}.
However, while much of the research in this field primarily focuses on defining and classifying toxic content, there is a limited exploration into the actual context in which such toxic content emerges \cite{markov-daelemans-2022-role,almerekhi2022provoke}. 
Furthermore, most of these studies predominantly focus on English \cite{vidgen2020directions}.

In this study, we investigate how social and political events trigger the start and spread of toxicity in online discussions across various languages. We further analyze the relationship between sentiment, emotion, and toxicity in relation to these world events. While toxicity scores highlight inappropriate conversations, sentiment analysis provides a broader spectrum. Combining toxicity,  sentiment, and emotion analyses allows for a more nuanced understanding of the conversational tone. For example, a comment with a negative sentiment or a strong emotion like anger but low toxicity could still contribute to a hostile environment, which might be overlooked if only toxicity is considered.
Our contributions can be summarized as follows:
\begin{itemize}
    \item We release 4.5 million comments organized as threads from 31K posts in six languages connected to major real-world events spanning four years from 2020 to 2023.
    \footnote{ All data and code are available on:\url{ https://github.com/cltl/grounding-toxicity} } 
    \item We explore the relationship between toxicity and real-world events in monolingual and cross-lingual settings.
     \item We explore the relationship between emotion, sentiment, and toxicity in relation to real-world events as a context across different languages, as well as over time.
\end{itemize}

In our data, we observed large variations in toxicity, sentiment, and emotion scores across different events and language communities, showing that toxicity is a complex phenomenon in which many different factors interact that still need to be investigated.

\section{Related Work}
We use toxic language as an umbrella term, similar to Sharma (2022) \cite{sharma2022detecting}, broadly comprising hate speech, offensive language, abusive language, propaganda, cyberbullying, and cyber-aggression. In this section, we provide an overview of studies that analyze one or more aspects of toxic language in social media settings from temporal, user and community perspectives.

\paragraph{Temporal and event analysis}
The study by Hiaeshutter (2022) \cite{hiaeshutter2022language} examines the relationship between significant social and political events in the U.S. during 2020 and 2021, and the increase in hostility within Reddit discussions. They study specifically how these events influenced language use in conversations. Their results reveal a strong link between key political events and a surge in hostility on Reddit. The study by Mall (2020) \cite{mall2020four} explores user behavior, specifically examining patterns of user toxicity over time. Their findings highlight that toxic users commonly alternate between making toxic and non-toxic comments.

\paragraph{User and community analysis} Kumar (2023) \cite{kumar2023understanding} provides an extensive study of the behavior of accounts on Reddit that post toxic content.  The study shows that even if abusive accounts constitute under 4\% of Reddit user base, they create 33\% of all the comments on Reddit. Similar work by Kumar (2018)\cite{kumar2018community} reveals a comparable trend, where a limited number of Reddit communities are responsible for a significant majority of negative interactions observed on the platform. 

In our work, we investigate how social and political events trigger the spread of toxicity in online discussions across various languages. We cover global and local real-world events across.

\section{Methodology}
Our goal is to collect conversations from Reddit across multiple languages that are grounded in the same major real-world events that are of interest to the community at large and that may give rise to both explicit and implicit toxic language. This data enables us to analyze 1) the discourse context of the toxicity of the conversation within comment threads, 2) the external world context in relation to the toxicity of the conversation, and 3) cross-lingual and cross-cultural factors in relation to the toxicity. To collect the data, we proceed in the following steps: 1) selection of the major real-world events and their descriptions in the target languages, 2) extraction of keywords for each event per language, 3) extraction of the Reddit submissions in each language related to these keywords, 4)  detection of the comments in the submissions with toxicity, negative sentiment, and negative emotions. In the following subsections, we explain the details of each step. 
\subsection{Event selection}
To enable a cross-language-community comparison, we target major world events. We selected major events between 2020 and 2023 from Wikipedia pages.
These events range from political events like the January 6 US Capitol attack and the COVID-19 pandemic to sports events like the 2022 FIFA World Cup. 

After identifying events, we utilize the corresponding Wikipedia page to gather the description of each event for the target language, which includes English, German, Dutch, Arabic, Turkish, and Spanish. These languages are selected because they represent major sub-communities in Europe. Each Wikipedia page includes links to versions of the content in various languages. We leverage these links to align the event descriptions with the target languages. We applied TF-IDF to the description of the events to extract keywords. We then use these keywords to collect Reddit posts related to the selected events.
\subsection{Data source and collection}
We use PRAW \footnote{ https://praw.readthedocs.io/en/stable/index.html}, the Reddit official Python package, to collect the data. We use two parameters for our search: the keywords we identify for each world event and the start date. Incorporating the start date in the query increases the probability that the submissions we obtain from our query are related to the specific target event. For instance, when dealing with the FIFA World Cup, utilizing only keywords might yield posts about past FIFA World Cups rather than the 2022 event. 
We then used PRAW to get lists of posts related to our target events, which resulted in 31 thousand posts. We used the submission IDs of these Reddit posts and collected all the comments and metadata under these submissions. We collected 4.5 million comments from 31 thousand posts.


\subsection{Toxic language detection}
We emphasize that our ultimate goal is to create a dataset across communities in which we can find subreddit threads with a high probability of containing toxic language. This should contain cases of not only explicit but also implicit toxic language. Because it is more difficult to find toxic language and most comments are not toxic \cite{vidgen2020directions}, we prefer a method that has a high recall of finding toxic comments so that we can further analyze the subthreads in which these occur. To decide on a high-recall method, we conduct a manual assessment of three methods to identify comment toxicity, focusing on those with the broadest applicability to our target languages. These methods include the Perspective API \cite{lees2022new}, a lexicon-based approach, and OpenAI's GPT-4 \cite{achiam2023gpt}.
\vspace{-10pt}
\subsubsection{Lexicon-based approach}
For the lexicon-based approach, we combine HurtLex  \cite{bassignana2018hurtlex}, MOL  \cite{vargas-etal-2021-contextual}, DALC  \cite{caselli2021dalc} and Hatebase  \cite{website:hatebase}
and build a binary classifier to score the toxicity of a comment. If at least one toxic word is present in a comment, we consider it toxic. Lexicon-based approaches have shown to be robust when detecting toxic words in cross-domain settings \cite{schouten2023cross} and can easily be extended to other languages or adapted in the future. Our merged lexicon has 4,316 English, 7,041 Dutch, 1,831 Arabic, 2,782 Turkish, 2,903 Spanish, and 2,851 German entries.
\subsubsection{GPT-4}
For GPT-4 \cite{achiam2023gpt}, we employ a simple zero-shot prompt to assign toxicity labels to a comment. We include a definition of a toxic comment in the prompt and prompt GPT-4 to classify comments as toxic or non-toxic. Our prompt is \textit{Review each comment and label it as toxic or non-toxic. To determine whether the comment is toxic if the comment falls into any of the following categories: hate speech, offensive language, abusive language, propaganda, cyberbullying, or cyber-aggression. If the comment aligns with any of these categories, label it as `Toxic' in the label column. If the comment does not fit any of these categories, label it as non-toxic}. 
\vspace{-10pt}
\subsubsection{Perspective API}
Perspective API is a Google-provided out-of-the-box toxicity classifier \cite{lees2022new}. The API takes a comment as input and produces a toxicity score between 0 and 1.  Based on the recommendation from the API documentation, we use a threshold value of 0.75 and consider a comment toxic if its toxicity score is higher than the threshold value.
\vspace{-10pt}
\subsubsection{Evaluation}
We evaluate the three methods by creating a validation set by experts. We randomly selected approximately 500 comments from each of the six target languages. We then select six experts who are also native speakers to annotate the sampled comments as toxic and neutral. We provided annotation guidelines with the definition of what kind of comments should be labeled as toxic. The experts also have the option to exclude comments that are code-mix or comments that are not in the target language. We exclude such comments from the evaluation. Our definition of toxic comment comprises hate speech, offensive language, abusive language, propaganda, cyberbullying, and cyber-aggression. We resolved annotators' questions and discrepancies through discussion. 
When evaluating the approaches for toxicity, we prioritize recall over precision because we want to maximize the probability that we find threads that exhibit explicit toxicity. Toxic comments are rare compared to non-toxic ones \cite{vidgen2020directions}; therefore, finding a single toxic comment flags a thread as potentially interesting, accepting some degree of false positives. 
The result of the evaluation of the three toxicity detection approaches is shown in 
Table~\ref{tab:aggregate_evaluation}. The lexicon-based approach shows a much better recall performance than Perspective-API and GPT-4. This is consistent across all the languages. Therefore, we used the lexicon-based approach to detect threads with likely toxic behavior.


\subsubsection{Sentiment and emotion lexicons}
In addition to scoring the comments and threads as potentially toxic, we also score comments for sentiment and emotion, for which we used the NRC emotion lexicon~\cite{Mohammad13} -- a manually annotated emotion lexicon in 100 languages. We use the 14,182 emotion words and their associations with eight emotions (anger, fear, anticipation, trust, surprise, sadness, joy, and disgust) and two sentiments (negative and positive) from the lexicon. We use a slightly different method for the sentiment and emotion analyses. Unlike toxicity, the presence of a single word may not be sufficient to classify a comment into a specific sentiment or emotion category. To determine the dominant sentiment and negative emotion value within a comment, we count the words in the comment that are listed in our negative sentiment lexicon. The sentiment of a comment is classified as negative if the count of words with negative polarity is greater than those with positive polarity. For the negative emotion score, we focus on words associated with anger, fear, sadness, and disgust and select the emotion represented by the highest word count. In cases where there is a tie in counts among these emotions, the comment is assigned multiple emotion classes. The lexicon we use is balanced across all six languages, containing 14,182 words for each.

\begin{table*}[ht]
\centering
\small
\begin{tabular}{@{}l@{\hspace{8pt}}l@{\hspace{8pt}}c@{\hspace{8pt}}c@{\hspace{8pt}}c@{\hspace{8pt}}c@{\hspace{8pt}}c@{\hspace{8pt}}c@{\hspace{8pt}}c@{\hspace{8pt}}c@{\hspace{8pt}}c@{\hspace{8pt}}|c@{\hspace{8pt}}c@{}}
\textbf{} & \textbf{} &
\multicolumn{3}{c}{\textbf{Lexical}} &
\multicolumn{3}{c}{\textbf{Perspective}} &
\multicolumn{3}{c}{\textbf{GPT-4}} \\
\cmidrule(lr){3-5} \cmidrule(lr){6-8} \cmidrule(lr){9-11}
& & \textbf{P} & \textbf{R} & \textbf{F1} &
\textbf{P} & \textbf{R} & \textbf{F1} &
\textbf{P} & \textbf{R} & \textbf{F1} & \textbf{Support}\\
\midrule
 & Non toxic  & .90 & .62 & .74 & .88 & .98 & .93 & .87 & .87 & .87 & 1315\\
& Toxic      & .17 & \underline{.53} & .25 & .35 & .08 & .13 & .08 & .08 & .08 & 190\\
& Macro avg  & .53 & .57 & .49 & .61 & .53 & .53 & .48 & .48 & .48 & 1505\\
\midrule
\rotatebox[origin=c]{90}{DE}  & Non toxic  & .97 & .46 & .63 & .96 & .93 & .95 & .94 & .94 & .94 & 240 \\
& Toxic      & .07 & .69 & .12 & .31 & .24 & .24 & .31 & .31 & .31 & 13\\
& Macro avg  & .52 & .58 & .37 & .58 & .62 & .59 & .47 & .47 & .47 & 253\\
\midrule
\rotatebox[origin=c]{90}{ES}        & Non toxic  & .88 & .46 & .61 & .82 & .99 & .88 & .86 & .86 & .86 & 178 \\
& Toxic      & .30 & .79 & .44 & .82 & .17 & .28 & .29 & .29 & .29 & 53 \\
& Macro avg  & .59 & .63 & .52 & .81 & .58 & .58 & .53 & .53 & .53 & 231 \\
\midrule
\rotatebox[origin=c]{90}{NL}        & Non toxic  & .97 & .38 & .55 & .94 & 1.00 & .97 & .96 & .96 & .96  & 252\\
& Toxic      & .08 & .81 & .14 & .00 & .00 & .00 & .12 & .12 & .12  & 16\\
& Macro avg  & .52 & .60 & .35 & .47 & .50 & .48 & .54 & .54 & .54  & 268\\
\midrule
\rotatebox[origin=c]{90}{AR}        & Non toxic  & .88 & .94 & .91 & .86 & 1.00 & .92 & .86 & .86 & .86  & 457\\
& Toxic      & .41 & .24 & .30 & .00 & .00 & .00 & .00 & .00 & .00  & 75\\
& Macro avg  & .65 & .59 & .61 & .43 & .50 & .46 & .43 & .43 & .43  & 532\\
\midrule
\rotatebox[origin=c]{90}{EN}        & Non toxic  &.86 & .51 & .64 & .85 & .95 & .90 & .84 & .84 & .84  & 188\\
& Toxic      & .16 & .55 & .25 & .17 & .06 & .09 & .06 & .06 & .06  & 33\\
& Macro avg  & .51 & .53 & .45 & .51 & .50 & .49 & .47 & .47 & .47  & 221\\
\midrule
\rotatebox[origin=c]{90}{TR}        & Non toxic & .69 & .57 & .62 & - & - & - & .60 & .87 & .71  & 180\\
& Toxic      & .49 & .61 & .54 & - & - & - & .37 & .12 & .18  & 120\\
& Macro avg  & .59 & .59 & .58 & - & - & - & .48 & .49 & .44  & 300\\
\bottomrule
\vspace{5pt}
\end{tabular}

\caption{Evaluation of Lexical-based approach, Perspective API and GPT-4. The first three rows show the aggregate result for all languages, followed by a language-specific breakdown. Here, we mark '-' since Perspective doesn't support Turkish. We also exclude Turkish from the aggregate result in the first three rows. The comments are lower than we originally sampled since we removed comments that are code-mix or comments that are not in the native language}
\label{tab:aggregate_evaluation}
\end{table*}

\section{Analysis}
In this section, we applied the above measurements to all the comments in our languages. These comments are grouped as threads related to world events and plotted in time. Grounding the comments to events and time allows us to apply a comparative analysis across events and language communities.

\begin{figure*}[ht]
  \centering
  \subfloat[Temporal distribution of comment across major world events. The distribution shows aggregate comments from all languages]
{\includegraphics[width=1.0\textwidth]{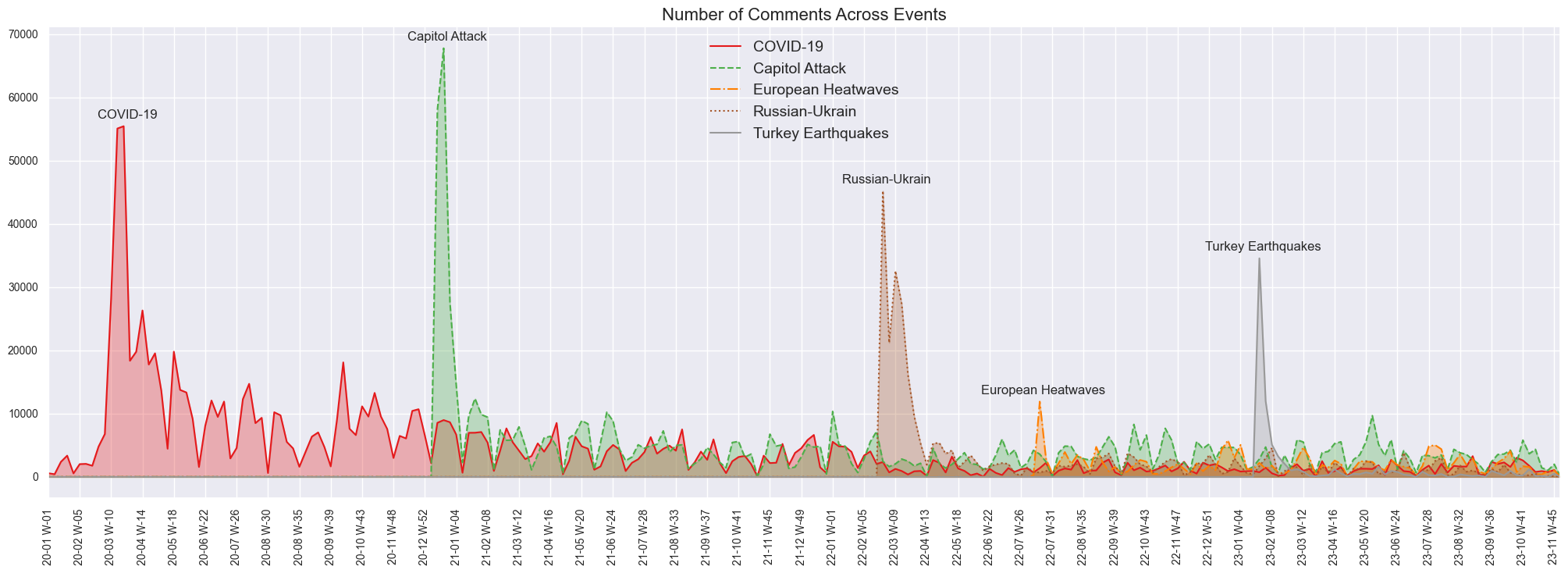}}\qquad
  \subfloat[Toxicity of comments for each event. We computed proportional toxicity by dividing the number of toxic comments by the total number of comments in a particular period.]{\includegraphics[width=1.0\textwidth]{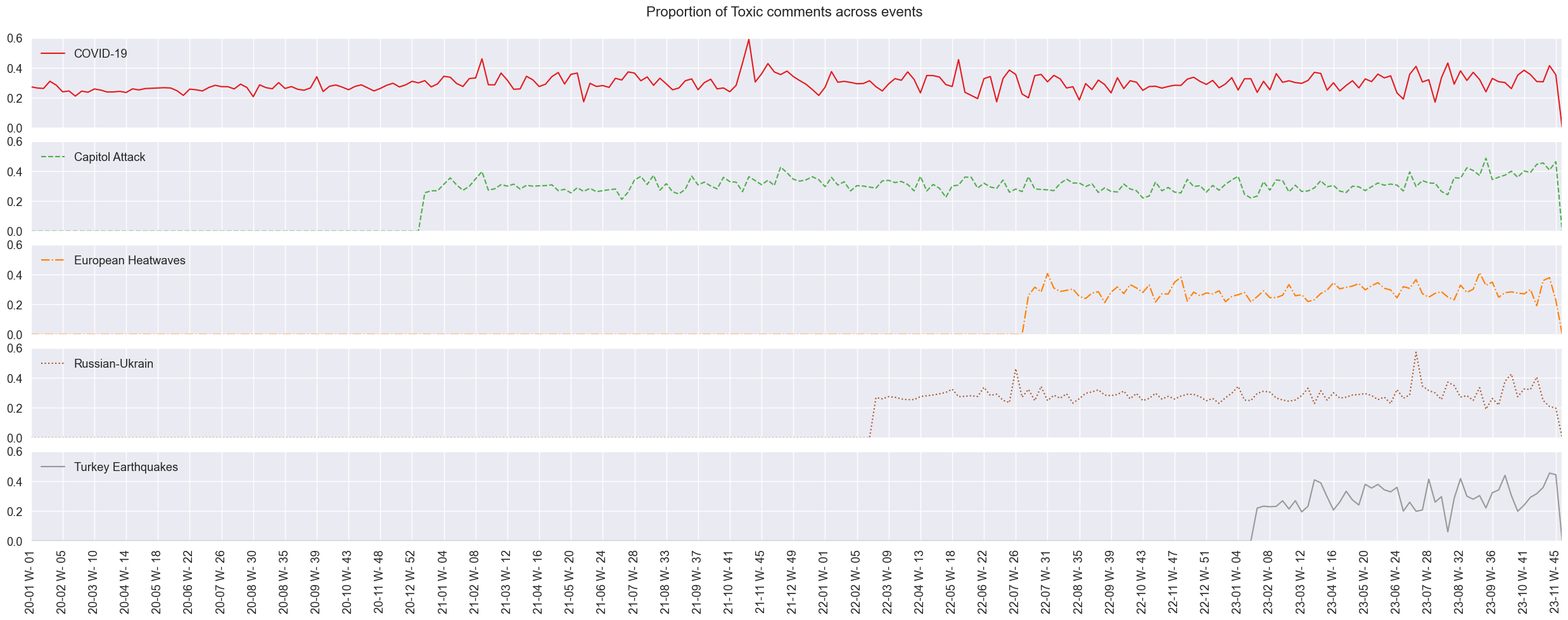} }\qquad
  \caption{Relationship between temporal event plot and the proportion of toxicity in conversation mentioning those events. The X-axis for the two plots is aligned to facilitate comparison. For visibility, we are only showing plots for five events.}
  \label{fig:count_and_toxicity}
\end{figure*}

\vspace{-20pt}

\subsection{Comment dynamics per event}
Figure \ref{fig:count_and_toxicity}(a) shows the temporal distribution of Reddit comments associated with various global events. The graph demonstrates the engagement intensity measured by the number of comments and duration over time, with spikes corresponding to specific events. The real-world events follow a well-defined pattern consisting of three distinct phases. Initially, there's the pre-event phase, characterized by minimal user discussions. This is followed by the peak phase, during which user engagement reaches its highest point, driven by real-time updates and widespread interest. Finally, there is a transition into the post-event phase, where activity decreases. Each event's temporal footprint is visible through the spikes in comment activity, which aligns with the actual dates of the events. Peaks in the graph correspond to the height of public discussion during these events, reflecting real-time reactions from the Reddit community. For instance, the COVID-19 peak signifies extensive discussions during the pandemic's onset in January and February 2020. Similarly, other events like the January 6 US Capitol Attack and the Fall of Kabul in 2021 show pronounced, albeit brief, increases in comment frequency. One possible explanation for this might be an observation reported by \cite{kumar2023understanding}, demonstrating that popular real-world events tend to engage active and previously inactive users, drawing them into discussions during these periods.

\subsection{Temporal analysis of toxicity}
In this section, we analyze the prevalence of toxicity in conversations in relation to major world events. We compute a normalized toxicity value by dividing the proportion of toxic comments over the total number of comments within a specific time period. Figure \ref{fig:count_and_toxicity}(b) shows the temporal pattern of toxicity, with each line representing a different major world event. The Y-axis quantifies the toxicity level, indicating the proportion of toxic comments during a given time frame, shown in the X-axis. We divide the X-axis in a week time block for easy comparison with Figure \ref{fig:count_and_toxicity}(a). In our analysis, we focus on two distinct scenarios. In the first scenario, we analyze if the increase in toxicity is observed during the peak period of an event's life cycle. In the second scenario, we analyze if we observe toxicity spikes during a non-peak phase of an event, possibly when the event is no longer recent.

\paragraph{Peak-related toxicity }
Contrary to our expectations, all the events show moderate or no significant peaks in toxicity close to the event outbreak, whereas we do see strong peaks in the volume of comments. On the other hand, toxicity peaks tend to occur later when the first outburst of comments has faded. A possible explanation could be that a larger community of Reddit users contributed in the beginning and that these users, being more representative of the average user, tend to use relatively less toxic language. After the attention to the event has faded, a smaller group of users remains, which apparently includes relatively more contributions of users that exhibit toxicity.

\begin{figure*}[ht]
  \centering
\includegraphics[width=1.0\textwidth]{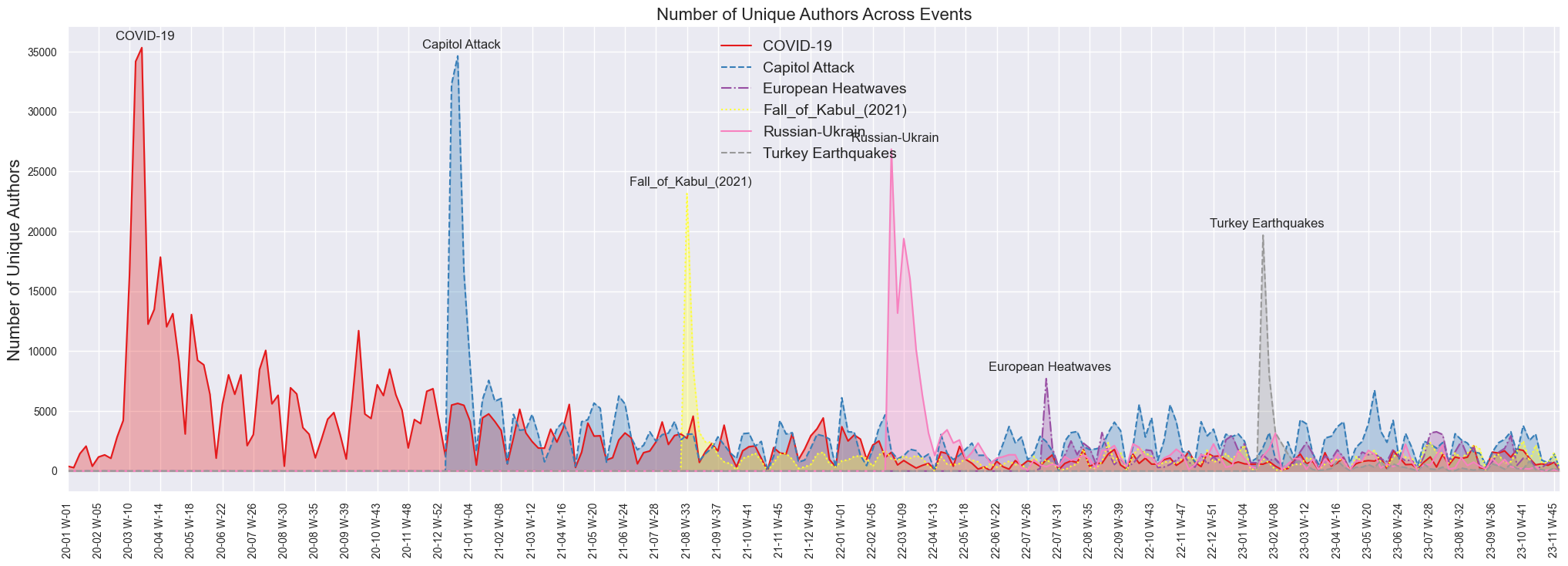}
  \caption{Temporal distribution of unique users. The spike in the number of comments in Figure \ref{fig:count_and_toxicity} (a) strongly correlates with an increased number of users engaged in commenting on a discussion about a particular event. For visibility, we are only showing plots for six events.}
  \label{fig:author_plot}
  \vspace{-20pt}
\end{figure*}

To control for the community factor, we quantified the number of unique users within the dataset and plotted this on the same timeline for the different events. The result is shown in Figure~\ref{fig:author_plot}. The volume of comments and the number of unique users show a very similar pattern over time, which is in line with our hypothesis that the larger community normalizes the proportion of toxicity.
\begin{table}[ht]
\centering
\begin{tabularx}{0.5\linewidth}{Xcc}  
\toprule
Event & Mean & Std-D \\
\midrule
COVID-19 & 0.297 & 0.056 \\
Capitol Attack & 0.23 & 0.143 \\
European Heatwaves & 0.1 & 0.14 \\
Kabul\_(2021) & 0.173 & 0.154 \\
Iran Protest & 0.088 & 0.138 \\
Israel Hamas & 0.008 & 0.045 \\
Russia-Ukraine & 0.128 & 0.147 \\
Turkey Earthquakes & 0.058 & 0.123 \\
\bottomrule 
\end{tabularx}  
\vspace{5pt}
\caption{Mean and standard deviation for toxicity of events.}
\vspace{-15pt}
\label{tab:events}
\end{table}
Table~\ref{tab:events} shows the mean and standard deviation for the toxicity density per event. 
The mean varies across the events, but the standard deviation is relatively low, which indicates that the toxicity remains more or less stable. The scores for COVID-19 and Israel-Hamas are expected: the former stretches a very long period, which evens out peaks, and the latter is too short to exhibit variation.

\begin{figure*}[ht]
  \centering
\includegraphics[width=1.0\textwidth]{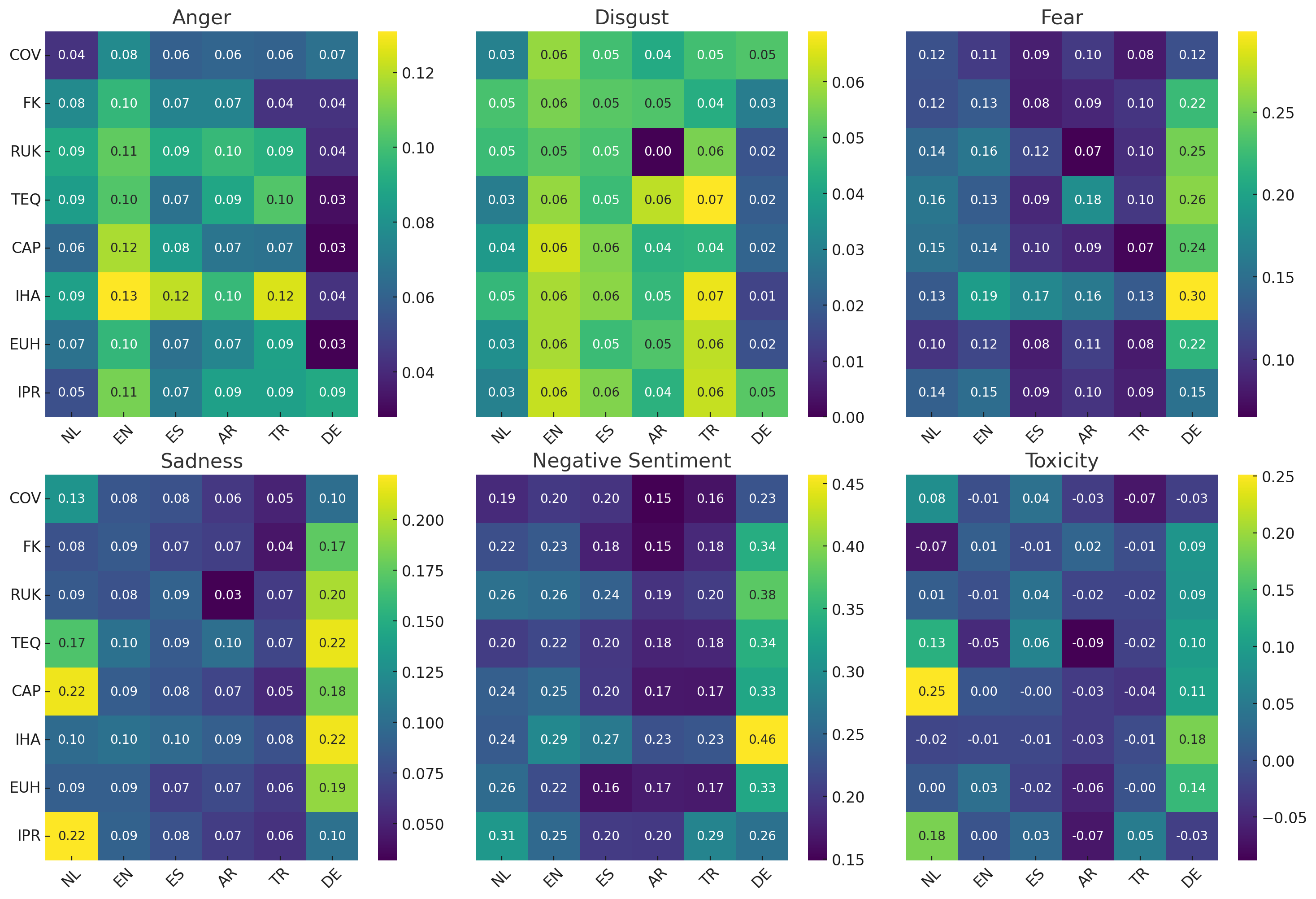}
  \caption{Heatmap for toxicity density, negative sentiment density, and negative emotions density. We use a shorter format for the event names to save space as follows- COVID-19:COV, Fall of Kabul:FK, Russia-Ukraine:RUK, Turkey Earthquak:TEQ, Capitol Attack:CAP, Israel Hamas:IHA, European Heatwaves:EUH, Iran Protest:IPR.
  }
  \vspace{-20pt}
  \label{fig:density_map}
\end{figure*}

\subsection{Cross-lingual analysis}
In this section, we analyze toxicity, sentiment, and emotions for each language to understand how these scores are influenced by language. We compute the density value for each score and plot them as heat maps. Figure \ref{fig:density_map} shows the density plot for toxicity, negative sentiment, and negative emotions (anger, disgust, fear, and sadness).
\vspace{-10pt}
\subsubsection{Toxicity analysis}
For this analysis, we compute the toxicity density by dividing the number of toxic comments in a particular event by the total number of comments in that event. We subtract the mean toxicity score from the density score to control for differences in toxic lexicon size across languages. 
When analyzing the toxicity across different languages and events, German and Dutch discussions stand out with higher average toxicity levels. This is particularly noticeable in Dutch discussions about the Capitol Attack, which exhibit the highest toxicity levels. Conversely, English and Turkish show lower average toxicity.
Event-wise, the Capitol Attack and the Iran Protests are associated with heightened toxicity. These two events, especially the Capitol Attack, show significant variability in toxicity levels across different languages. On the other hand, events like COVID-19 and the Fall of Kabul are characterized by lower average toxicity, suggesting more neutral conversations surrounding these topics. While the highest toxicity is observed in Dutch discussions during the Capitol Attack, the lowest is seen in Arabic during the Turkey Earthquake. The standard deviation in toxicity levels, particularly high for German, indicates a wide range of reactions expressed across different events. This variability, coupled with the specific peaks in toxicity, highlights the complex relationship between real-world events and the cultural and linguistic nuances that shape online discussions surrounding these events.
\vspace{-10pt}
\subsubsection{Sentiment analysis}
Similar to the toxicity density, we compute a normalized negative sentiment density score by dividing the number of comments with negative sentiment by the total number of comments for each event. Since the size of the sentiment lexicons across languages is equal, we do not subtract the mean value here.
A notable trend is the consistently higher negative sentiment in German conversation across all events, with a peak during the Israel-Hamas and the Russia-Ukraine conflicts, indicating a strong emotional response to this event within the German-speaking community. Conversely, Arabic conversations consistently exhibit the lowest negative sentiment, while English and Turkish show relatively consistent sentiment levels. The result reflects how linguistic communities' online responses to global crises vary. 
\vspace{-10pt}
\subsubsection{Emotion analysis}
In the emotion analysis, we compute the density from four negative emotions: anger, fear, sadness, or disgust. 
In the anger heatmap, conversations in English consistently show higher anger scores, most prominently during the Capitol Attack. In contrast, German conversations exhibit significantly lower anger across all events, a stark contrast to their higher negative sentiment. Arabic conversations also show higher scores in certain events like the Fall of Kabul and the Russia-Ukraine conflict. In the disgust heatmap, similar to the anger heatmap, we observe equally higher levels of disgust expressed in English in most events. Spanish conversations also show similar patterns to those of English. German, in contrast, consistently exhibits the lowest levels of disgust across all events.

We observe a similarity between the trend in the fear heatmap and the negative sentiment heatmap. German conversations show a notably high overall score, especially prominent during the periods of the Russia-Ukraine conflict and the Turkey Earthquake.
In contrast, Arabic conversations generally show lower levels of fear, except for a notable increase during the Turkish Earthquake. Overall, German conversations consistently displayed higher levels of negative sentiment and sadness across all events, yet interestingly, their levels of anger and disgust were notably lower. Comparatively, English conversations frequently exhibited higher anger and disgust, particularly during events like the Capitol Attack and the Turkey Earthquake. On the other hand, Arabic conversations generally showed lower levels of negative sentiment, anger, and sadness, with certain exceptions indicating event-specific responses. The stronger toxicity of the Dutch community is not reflected by equally strong negative sentiments and emotions when compared to the other languages, except for sadness. 
\vspace{-10pt}
\subsubsection{Correlation analysis}
We further applied the Pearson correlation analysis of toxicity, emotion, and sentiment within and across languages. There is a general trend of positive correlations among individual languages, though the strength varies. Fear, sadness, and negative sentiment exhibit strong relationships, suggesting a consistent link between these emotions within languages. For instance, a correlation of 0.72  between fear and sadness within languages. Across languages, this strong correlation pattern holds. However, the correlation strengths differ within and across languages, suggesting cultural variation in how emotions and sentiments are expressed and perceived. This nuanced analysis underscores the complexity of emotional responses across different cultural and language-community contexts while pointing to some universal patterns in emotional expression. All our observations so far depend on the correctness of the instruments to measure these properties. We realize that this analysis is a rough indication pointing to interesting data for further analysis to be carried out in the future.
\section{Conclusion}
In this study, we examine the changes in toxicity, sentiment, and emotions within Reddit discussions during major social and political events, analyzing these aspects across various languages. Our findings show three key points. First, we found that toxicity in comments usually increases later, not right at the start of events. One possible reason for this might be that the first group of commenters are more diverse and less likely to use toxic language. As the event becomes less popular, the remaining users tend to show more toxic behavior. Second, we observe a lot of variation when we looked at toxicity, sentiment, and emotions across different language communities. This shows that language communities respond differently to global events, and it is essential to consider cultural context when studying online interactions. Lastly, we found a positive correlation between toxicity, sentiment, and emotion in each language, but the strength of this link varies. Emotions like fear and sadness often go together with negative sentiments, showing a consistent pattern in how these emotions are connected across languages. It is evident that the reliability of our observations relies on the assumption that the instruments utilized for quantifying toxicity, sentiment, and emotion are accurate.
However, we will use this data in further research to analyze the implicit and explicit toxic reasoning towards specific targets and in relation to the events as contexts. Our preliminary analysis shows the potential to find such patterns. The fact that the data, within limits, is grounded in real-world events provides a strong basis for cross-lingual and cross-community comparison in the future.

\section*{Limitations}
We identified some limitations in our work. First, the lexicons we use for detecting toxicity, sentiment, and emotion might have differences in quality across each language. These might introduce
bias and result in less accurate results across languages. Second, we extracted comments using keywords and temporal constraints to link these with real-world events. This does not necessarily mean that the submissions and comments actually discuss these events as a topic. The actual context of the online discussion could still vary and be different. Further topic analysis of the posts and comments should clarify this in future work.

\section*{Acknowledgements}
The research was supported by Huawei Finland through the DreamsLab project. All content represented the opinions of the authors,which were not necessarily shared or endorsed by their respective employers and/or sponsors.

\bibliography{current}
\bibliographystyle{bst_file/splncs04}

\end{document}